\documentclass[11pt]{article}

\usepackage{acl}

\usepackage{times}
\usepackage{latexsym}

\usepackage[T1]{fontenc}

\usepackage[utf8]{inputenc}

\usepackage{microtype}

\usepackage{inconsolata}

\usepackage{graphicx}
\usepackage{booktabs}
\usepackage{float}
\usepackage{amsmath}
\usepackage{soul} 
\newcommand{\colorul}[2]{\setulcolor{#1}\ul{#2}}
\setul{1.5pt}{0.3ex} 

\definecolor{female}{RGB}{132,33,244}
\definecolor{male}{RGB}{237,117,46}
\definecolor{agricultural}{RGB}{54,33,95}
\definecolor{manufacturing}{RGB}{71,97,214}
\definecolor{construction}{RGB}{44,183,240}
\definecolor{sciences}{RGB}{27,228,183}
\definecolor{logistics}{RGB}{50,242,152}
\definecolor{security}{RGB}{92,252,111}
\definecolor{cleaning}{RGB}{114,254,95}
\definecolor{tourism}{RGB}{156,255,64}
\definecolor{trade}{RGB}{136,255,79}
\definecolor{management}{RGB}{186,246,53}
\definecolor{office}{RGB}{200,239,53}
\definecolor{HR}{RGB}{213,231,54}
\definecolor{finance}{RGB}{235,209,59}
\definecolor{law}{RGB}{244,199,58}
\definecolor{healthcare}{RGB}{253,141,39}
\definecolor{education}{RGB}{202,42,4}
\definecolor{social_work}{RGB}{188,32,2}
\definecolor{design}{RGB}{140,12,5}
\definecolor{journalism}{RGB}{158,17,1}
\definecolor{media}{RGB}{121,4,4}

\usepackage{tikz}
\usetikzlibrary{shapes.geometric, arrows,decorations.pathreplacing}
\tikzstyle{approach} = [rectangle,text width=2.5cm, minimum height=2.5cm, text centered, draw=black, thick]
\tikzstyle{arrow} = [thick,-]
\tikzstyle{start} = [rectangle,text width=4cm, minimum height=1cm, text centered, draw=black, thick]
\tikzstyle{small} = [rectangle,text width=2.5cm, minimum height=1.5cm, text centered, draw=black, thick]
\tikzstyle{smaller} = [rectangle,text width=2cm, minimum height=1.5cm, text centered, draw=black, thick]
\tikzstyle{big} = [rectangle,text width=4cm, minimum height=1.5cm, text centered, draw=black, thick]
\tikzstyle{nope} = [rectangle, text width=2.5cm, minimum height=1.5cm, text centered, draw=white, thick]
\tikzstyle{word} = [rectangle, text width=4cm, minimum height=1cm, draw=white, thick]
\tikzstyle{sentence} = [rectangle, text width=10cm, minimum height=1cm, draw=white, thick]
\tikzstyle{ngram} = [rectangle, text width=2.9cm, minimum height=0.5cm, draw=black, thick]
\tikzstyle{ngram1} = [rectangle, text width=5.5cm, minimum height=0.5cm, draw=black, thick]
\tikzstyle{word2} = [rectangle, minimum height=1cm, draw=white, thick]
\tikzstyle{sen} = [rectangle, minimum height=1cm, draw=white, thick]

\title{Are All Spanish Doctors Male? \\Evaluating Gender Bias in German Machine Translation}

\author{Michelle Kappl \\
  Technische Universität Berlin\\
  \texttt{michelle.kappl@tu-berlin.de}}

\begin{document}
\maketitle
\begin{abstract}

We present WinoMTDE, a new gender bias evaluation test set designed to assess occupational stereotyping and underrepresentation in German machine translation (MT) systems. Building on the automatic evaluation method introduced by \citet{stanovsky-etal-2019-evaluating}, we extend the approach to German, a language with grammatical gender. The WinoMTDE dataset comprises 288 German sentences that are balanced in regard to gender, as well as stereotype, which was annotated using German labor statistics. We conduct a large-scale evaluation of five widely used MT systems and a large language model. Our results reveal persistent bias in most models, with the LLM outperforming traditional systems. The dataset and evaluation code are publicly available at \url{https://github.com/michellekappl/mt_gender_german}.

\end{abstract}

\section{Introduction}

In a globalized world, millions rely on machine translation (MT) systems to break language barriers in medicine, business, and diplomacy every day \citep{vieira_understanding_2021}. However, when these systems fail, the consequences can be severe \citep{Canfora_Ottmann_2020}. The results of a study conducted by \citet{patil_use_2014} show Google
Translate incorrectly translating the phrase \textit{``Your child is fitting''} (which denotes a child
having seizures) into the Swahili equivalent of \textit{``Your child is dead''}. While translation errors in medical contexts can lead to life-threatening misunderstandings, it is not the only domain where MT systems can fail. Another example is depicted in Figure~\ref{fig:gender_bias_example}, where Google Translate mistranslates a sentence from German to Spanish.

\begin{figure}[h]
  \centering
  \resizebox{\columnwidth}{!}{%
  \begin{tikzpicture}
      \node (sen) [sen] {\textbf{\textcolor{female}{Die Managerin$_f$}} feuerte \textbf{\textcolor{male}{den Reiniger$_m$}}, weil sie wütend war.\footnotemark};
      \node (trans) [sen, below of = sen, yshift = 0cm] {\textbf{\textcolor{male}{El gerente$_m$}} despidió a \textbf{\textcolor{female}{la limpiadora$_f$}} porque estaba enojada.};
      \begin{scope}[transform canvas={xshift=0cm}]
          \draw [->] (0,-0.25) -- ++(0,-0.4cm);
      \end{scope}
      \begin{scope}[transform canvas={xshift=-4cm}]
          \draw [->] (0,-0.25) -- ++(0,-0.4cm);
      \end{scope}
  \end{tikzpicture}}
  \caption{Example of gender bias in German Machine Translation by Google Translate, where occupational stereotypes are reinforced.}
  \label{fig:gender_bias_example}
\end{figure}
\footnotetext{Translation: The manager$_f$ fired the cleaner$_m$, because she was mad.}

In this case, the German noun \textit{Die Managerin}, explicitly marked as female, was mistranslated into the masculine Spanish term \textit{El gerente}. Despite clear grammatical markers indicating the subject's gender, the MT system defaulted to a male translation, thereby producing a flawed translation. These phenomena are referred to as gender bias in MT and of rising concern in the field of natural language processing \citep{savoldi_gender_2021,costa-jussa_analysis_2019,blodgett-etal-2020-language}.

\paragraph{Bias Statement \citep{blodgett-etal-2020-language}.}
Gender-biased translations reinforce societal assumptions about the roles and abilities of different genders \citep{vervecken_yes_2015,sterling_confidence_2020}. If MT models systematically misrepresent female subjects and reinforce stereotypical gender roles in occupational contexts, they contribute to the invisibility of women in professions traditionally dominated by men \citep{horvath_does_2016}. Research has shown that children are particularly susceptible to such biases, which can shape their perceptions of career difficulty, prestige, and self-efficacy \citep{vervecken_yes_2015}. Furthermore, \citet{vervecken_yes_2015} found that using pair-forms (e.g., \textit{``Feuerwehrmänner und Feuerwehrfrauen''} for male and female firefighters) instead of male generics increases children's confidence in pursuing non-traditional careers. Studies also highlight a correlation between women's self-efficacy in STEM (Science, Technology, Engineering, and Mathematics) occupations and the persistent gender pay gap \citep{sterling_confidence_2020}.
\paragraph{Contribution.}
To address these issues and minimize potential harm, it is crucial to deepen our understanding of gender bias in MT. Prior research has primarily focused on English MT models, with \citet{stanovsky-etal-2019-evaluating} conducting the first large-scale evaluation on this topic. This work aims to bridge existing research gaps by introducing a German gender bias evaluation testset (GBET), WinoMTDE, which extends the proposed automatic evaluation method developed by \citet{stanovsky-etal-2019-evaluating} to German. The dataset is designed to evaluate occupational stereotyping and gender bias in German MT and therefore enabled us to do a systematic analysis of five widely used MT systems namely Google Translate, Microsoft Translator, Amazon Translate, DeepL, and SYSTRAN. In addition to these traditional MT systems, we also assess GPT-4o-mini, as large language models are increasingly integrated into everyday applications and frequently used for translation tasks \citep{Chan_Tang_2024}. These models were evaluated on their ability to correctly translate sentences from German to seven target languages that heavily exhibit gender in their grammatical structure: French, Italian, Spanish, Ukrainian, Russian, Arabic, and Hebrew. Unlike English, German uses explicit grammatical gender markers, which should, in theory, reduce ambiguity when translating into other gendered languages. One might expect MT systems to produce more accurate and gender-consistent translations due to these grammatical cues. However, despite the availability of such markers, our findings reveal that gender bias persists in most models. This indicates that the problem stems from systemic biases within model architectures and training data rather than source-language ambiguity.
\paragraph{Related Work.}
\citet{stanovsky-etal-2019-evaluating} conducted the first large-scale evaluation of gender bias in English MT systems. They introduced the \textit{WinoMT} GBET, which is based on two corpora of sentences following the Winograd schema \citep{levesque_winograd_2012}, namely \textit{Winogender} \citep{rudinger_gender_2018} and \textit{WinoBias} \citep{zhao_gender_2018}. In their evaluation, they found that all tested MT systems exhibited significant stereotypical and gender bias.

\section{Methodology}
In this section we introduce the WinoMTDE dataset, discuss the evaluation pipeline, and outline the used metrics.

\subsection{WinoMTDE}
We introduce the WinoMTDE\footnote{available at \url{https://github.com/michellekappl/mt_gender_german}} dataset, a German GBET which is a translated subset of WinoMT by \citet{stanovsky-etal-2019-evaluating}. The dataset consists of 288 German sentences structured according to the Winograd schema (see Figure \ref{fig:gender_bias_example}), where each sentence consists of a clearly gendered subject of interest (e.g. \textit{Die Managerin}) in the main clause, as well as another subject of opposite gender (e.g. \textit{der Reiniger}). A pronoun (e.g., \textit{sie}) in a dependent clause refers to the subject of interest. WinoMTDE currently includes only binary-gendered terms and pronouns. It does not account for non-binary pronouns or neutral occupational terms. Each sentence is annotated with:
\begin{itemize}
  \item The subject's \textbf{gender} (male or female).
  \item Its \textbf{position} in the sentence.
  \item The \textbf{stereotype alignment}, i.e. if the occupation is pro- or anti-stereotypical.
\end{itemize}
The dataset is balanced in regard to gender, containing an equal number of female and male-gendered subjects of interest (144 each). \citet{stanovsky-etal-2019-evaluating} used statistics from the U.S. Department of Labor to split WinoMT into equal parts pro- and anti-stereotypical instances. This is used for further evaluating each MT model regarding stereotypical
gender bias. For the WinoMTDE testset to better reflect the German society, statistics from the German Department of Labor (Bundesagentur für Arbeit) were used. Each occupation of the WinoMTDE set was classified according to the \textit{``German Classifications of Occupations 2010 - Revised Version 2020''} \citep{statistik_der_bundesagentur_fur_arbeit_klassifikation_2020}. This classification can be found in the appendix (see \ref{app:occupation_statistics}). By considering the gender distribution of each classified occupation, the stereotypical
gender (defined as more than 50\%) associated with each occupation was determined. For example, the female occupation \textit{Managerin} falls under the category
\textit{``711 - Geschäftsführung und Vorstand''} (managing and board members). Given that 77\% of individuals working in this field are male, the sentence containing \textit{Managerin} is classified as anti-stereotypical.
These subsets, called WinoMTDE$_{anti}$ and WinoMTDE$_{pro}$, contain 121 instances each, therefore making WinoMTDE balanced in regard to stereotype as well. The reduction in size stems from nouns that can not be classified, such as \textit{PatientIn} (patient) or \textit{BesucherIn} (visitor).
\subsection{Evaluation Pipeline}
  \begin{figure}[h]
      \centering
      \includegraphics[width=\columnwidth, height=0.2\textheight]{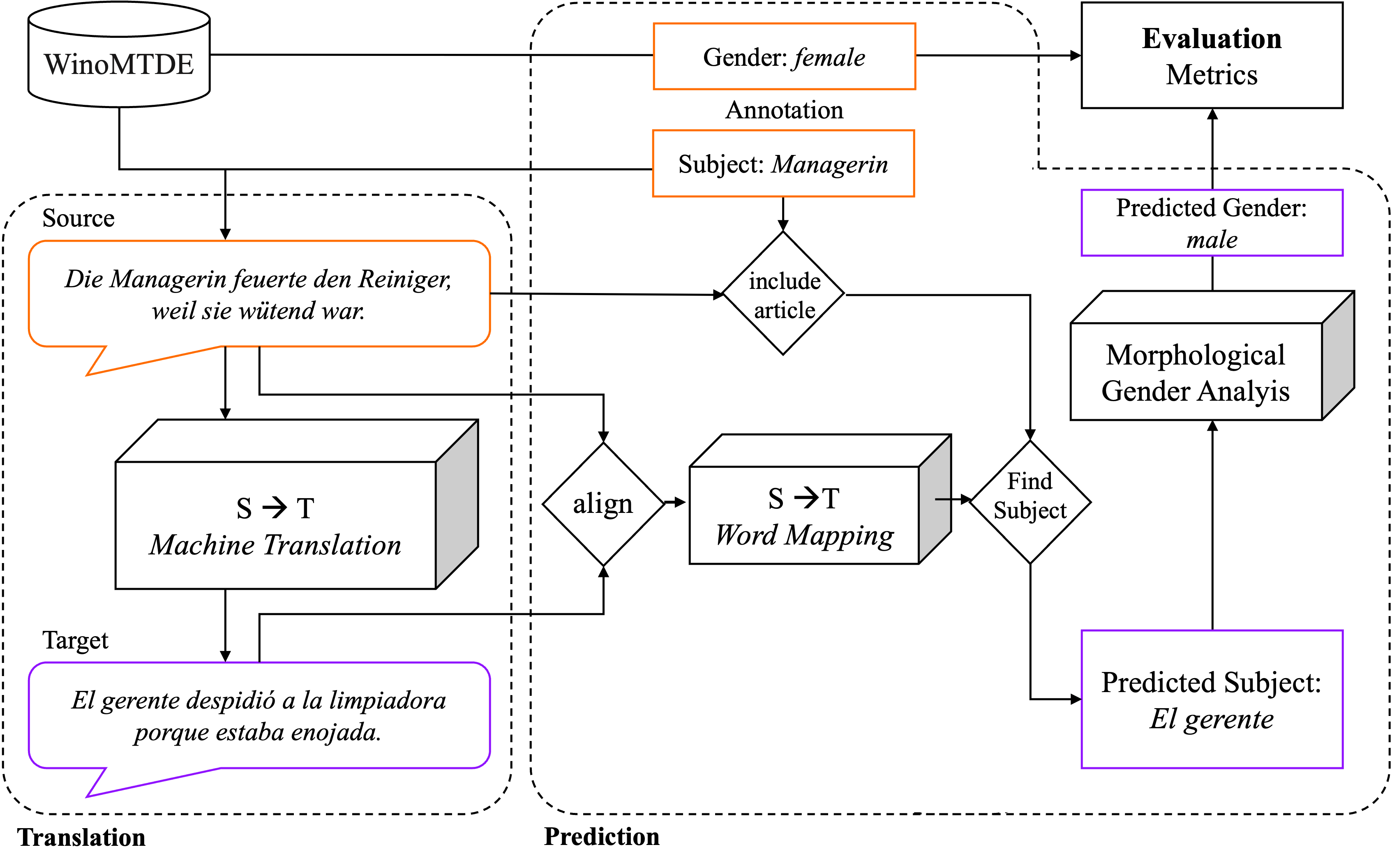}
      \caption[Evaluation Pipeline]{Evaluation pipeline \citep{citekey}. The German ground truth is indicated by \textcolor{male}{orange} and the translation by the MT model and the corresponding gender and subject predictions are indicated by \textcolor{female}{violet}.}
      \label{fig:pipeline}
  \end{figure}

  The evaluation pipeline (Figure \ref{fig:pipeline}) based on the work of \citet{stanovsky-etal-2019-evaluating} evaluates translations from German into seven target languages, discussed in \nameref{par:languages}. The pipeline can be divided into three main steps: \textbf{translation}, \textbf{prediction}, and \textbf{evaluation}.
  \paragraph{Translation.}
  As illustrated in Figure \ref{fig:pipeline}, the pipeline is designed to translate each sentence $S$ from the German WinoMTDE testset into a target language, thus producing a corresponding translation $T$ using a selected MT model $M$.
  \begin{table*}
    \resizebox{\textwidth}{!}{%
    \begin{tabular}{cccccccccccccccccccccccccc}
    \toprule
    \multicolumn{1}{c}{} & \multicolumn{3}{c}{\textbf{Google Translate}} & \multicolumn{1}{c}{} & \multicolumn{3}{c}{\textbf{Microsoft Translator}} & \multicolumn{1}{c}{} & \multicolumn{3}{c}{\textbf{Amazon Translate}} & \multicolumn{1}{c}{} &  \multicolumn{3}{c}{\textbf{SYSTRAN}} & \multicolumn{1}{c}{} &  \multicolumn{3}{c}{\textbf{DeepL}} & \multicolumn{1}{c}{} &  \multicolumn{3}{c}{\textbf{GPT-4o-mini}} \\
  
    \cmidrule(lr){2-4} \cmidrule(lr){6-8} \cmidrule(lr){10-12} \cmidrule(lr){14-16} \cmidrule(lr){18-20} \cmidrule(lr){22-24}
  
    \textbf{Languages} & {\textsc{Acc}} & {$\Delta_G$} & {$\Delta_{S}$} && {\textsc{Acc}} & {$\Delta_G$} & {$\Delta_{S}$} && {\textsc{Acc}} & {$\Delta_G$} & {$\Delta_{S'}$} && {\textsc{Acc}} & {$\Delta_G$} & {$\Delta_{S}$} && {\textsc{Acc}} & {$\Delta_G$} & {$\Delta_{S}$} && {\textsc{Acc}} & {$\Delta_G$} & {$\Delta_{S}$}\\
    \midrule
  
    \textit{DE$\rightarrow$ES} & \underline{66.8} & 11.9 & 15.6 && 62.0 & 16.8 & 11.1 &&  \underline{72.7} & 5.2 & 6.8 && \underline{94.1} & 0.1 & 6.6 && 83.1 & 6.4 & 5.6 && \textbf{\underline{95.8}} & 1.7 & -0.8 \\
    \textit{DE$\rightarrow$FR} & 64.2 & 12.1 & 16.2 && \underline{69.2} & 6.2 & 20.9 && 68.0 & 5.7 & 24.5 && 80.6 & 1.5 & -2.7 && \underline{83.3} & 0.4 & -2.3 && \textbf{89.4} & 0.8 & 1.3 \\
    \textit{DE$\rightarrow$IT} & 52.0 & 26.2 & 14.2 && 51.8 & 31.8 & 14.4 && 58.9 & 16.8 & 13.2 && 70.9 & 7.7 & 6.0 && 61.9 & 15.8 & 13.7 && \textbf{75.5 }& 4.4 & 2.3 \\
    \midrule
    \textit{DE$\rightarrow$UK} & 46.5 & 14.7 & 11.6 && 48.2 & 18.8 & 4.0 && 41.4 & 27.4 & 8.0 && 38.2 & 27.4 & -8.2 && 54.7 & 8.2 & 11.7 && \textbf{69.6} & 6.1 & -2.4 \\
    \textit{DE$\rightarrow$RU} & 42.7 & 19.4 & 6.4 && 46.4 & 15.6 & 8.2 && 47.3 & 15.6 & 8.2 && 37.0 & 22.5 & 6.9 && 42.3 & 15.5 & -3.0 && \textbf{55.4} & 6.3 & -15.5 \\
    \midrule
    \textit{DE$\rightarrow$AR} & 55.2 & 18.3 & 9.0 && 54.0 & 20.8 & 9.2 && 59.2 & 15.3 & 7.5 && 51.5 & 24.3 & 10.9 && - & - & - && \textbf{83.3} & 0.5 & 4.9 \\
    \textit{DE$\rightarrow$HE} & 64.5 & 3.8 & 17.5 && 65.4 & 1.9 & 20.9 && 60.3 & 10.0 & 18.7 && 44.6 & 16.1 & 18.4 && - & - & - && \textbf{78.1} & -1.5 & 5.4 \\
    \bottomrule
    \end{tabular}}
    \caption[Results of this Evaluation]{Results of this evaluation for all language pairs. Languages are grouped into their respective language families: Romance, Slavic, and Semitic. The highest accuracy result for each language pair (row-wise) is highlighted in bold, while the best result for each MT model (column-wise) is underlined. DeepL is unable to translate German to either Arabic or Hebrew, which is why the corresponding cells are left empty.}\label{tab:results}
  \end{table*}
  \paragraph{Prediction.} 
  Using \textit{fast-align} the source and target sentence get mapped to one another. It is a word alignment tool that was developed by \citet{dyer_simple_2013} and produces a word alignment in the "Pharao format". For each word index in $S$ \textit{fast-align} finds the corresponding word index in $T$.  This means that the word, that is our subject of interest in $S$ is aligned with the corresponding word in $T$. Furthermore, especially in the Romance languages, where each noun has a gendered noun determiner, the gender is often clearly encoded in the articles. To improve prediction quality the subject of interest as well as the corresponding article are used for the morphological analysis. Language-specific tools   like \textit{spaCy} (for Romance languages), \textit{pymorphy2} (for Slavic languages), and the morphological analyzer by \citet{adler_unsupervised_2006} (for Hebrew) determine the gender of the nouns. For Arabic, gender is inferred using the ta marbuta character, a marker of femininity. If it is not possible to determine the gender of a word, it is marked as unknown. Furthermore, gender-neutral terms, such as the Spanish word \textit{estudiante} (student, no specified gender) are annotated as neutral.
  Using the predicted gender information on the translated subject of interest different metrics are calculated to evaluate the MT model $M$.
  \paragraph{Evaluation.}
  The evaluation is based on the following metrics.
  \begin{description}
    \item[Accuracy.] For each model $M$ the general accuracy is calculated and denotes the percentage of instances where the ground truth gender (annotated in WinoMTDE) matches the predicted gender. It is calculated as follows:
    \begin{align*}
      \text{\textsc{acc}} = \frac{\text{total number of correct predictions}}{\text{total number of predictions}}
    \end{align*}
    \item[Gender-based F1-score gap $\Delta_G$.] The F1-score is a metric that combines precision and recall. Precision is defined as the ratio between correct predictions and the total number of predictions. Recall on the other hand is the ratio between correct predictions and the total number of instances. Both of these metrics are calculated using the WinoMTDE set as the ground truth and with the following formulas, where the gender $g$ is either male Using this the respective F1-Scores can be calculated as follows:
    \begin{align*}
        \text{\textsc{F1-score}}_g &= 2 \cdot \frac{\text{Precision}_g \cdot \text{Recall}_g}{\text{Precision}_g + \text{Recall}_g}
    \end{align*}
    After calculating both the male and the female \textsc{F1-score}, $\Delta_G$ is defined by following formula:
    \begin{align*}
        \Delta_{G} = \text{\textsc{F1-score}}_m - \text{\textsc{F1-score}}_f
    \end{align*}
    \item[Stereotype-based performance gap $\Delta_{S}$.]
    \citet{stanovsky-etal-2019-evaluating} defines $\Delta_S$ as the "difference in performance (F1-score)\footnote{It is important to note that even though the paper states that it utilizes the F1-score (although no formula is given) the actual calculation within the code published on GitHub is done using Accuracy.} between stereotypical and non-stereotypical gender role assignments". In contrast to the metrics discussed previously, it utilizes the subsets of WinoMTDE that are classified as stereotypical (WinoMTDE$_{pro}$) and anti-stereotypical (WinoMTDE$_{anti}$). $\Delta_S$ is calculated as follows:
    \begin{align*}
        \Delta_{S} = \text{\textsc{Acc}}_{pro} - \text{\textsc{Acc}}_{anti}
    \end{align*}
\end{description}

\section{Experimental Setup}
Using the evaluation pipeline, we evaluate six MT models on seven languages.
\paragraph{MT Models.} The original paper by \citet{stanovsky-etal-2019-evaluating} evaluated five commercial MT systems, namely Google Translate (GT), Microsoft Translator (Micr. T), Amazon Translate (AT), and SYSTRAN (S). In addition to these models, we also evaluate DeepL (D) and GPT-4o-mini (4o-m). The models were selected based on their popularity, availability, and the comparability of the results with the original study. Most of these models are neural machine translation systems, except for SYSTRAN, which is a hybrid system combining rule-based and statistical MT and GPT-4o-mini, a large language model.
\paragraph{Languages.}\label{par:languages} The models are evaluated on their ability to translate German sentences into seven target languages: Hebrew (HE), Arabic (AR), Spanish (ES), French (FR), Italian (IT), Russian (RU), and Ukrainian (UK). These languages were selected based on their gendered grammatical structure and different language families.

\section{Results}
The main results of this evaluation are presented in Table \ref{tab:results}, highlighting the performance of each model across accuracy, gender-based F1-score gaps ($\Delta_G$), and stereotype-based performance gaps ($\Delta_S$). 
\begin{figure*}[h]
  \centering
  \includegraphics[width=\textwidth]{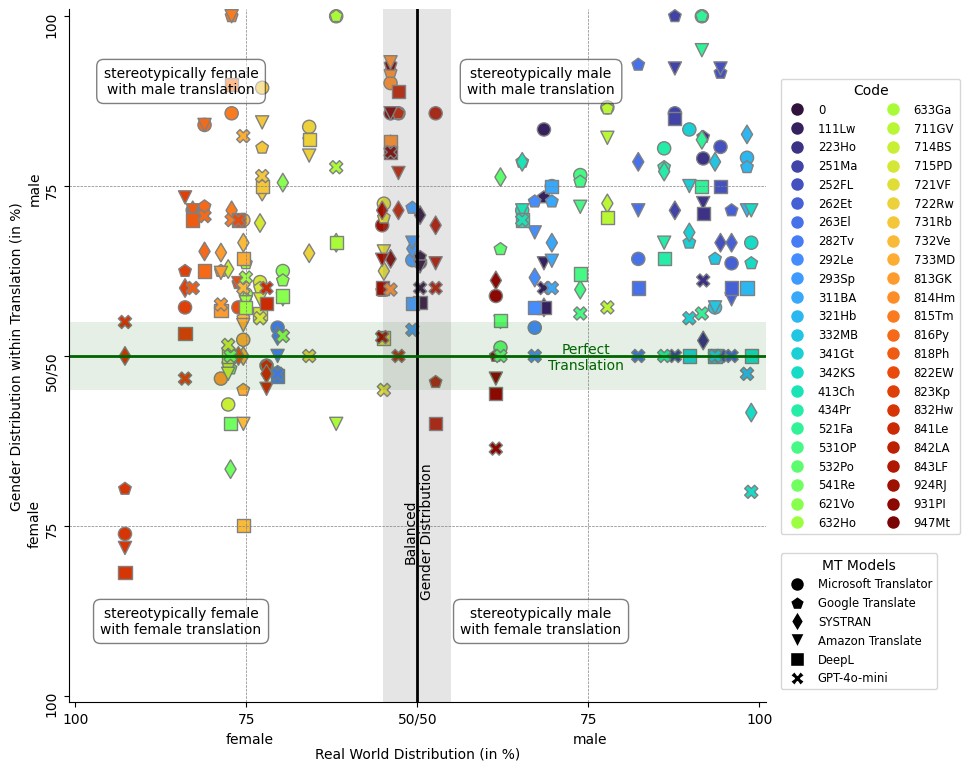}
  \caption[Gender Distribution for all Occupation Groups and Models]{Gender predictions for each occupation group across all languages and MT models were aggregated and visualized. Colors represent professional categories: blue hues for \colorul{agricultural}{agricultural}, \colorul{manufacturing}{manufacturing}, and \colorul{construction}{construction}; turquoise for \colorul{sciences}{sciences}, \colorul{logistics}{logistics}, and \colorul{security}{security}; green for \colorul{cleaning}{cleaning}, \colorul{tourism}{tourism}, and \colorul{trade}{trade}; greenish-yellow for \colorul{management}{management}, \colorul{office}{office}, and \colorul{HR}{HR}; yellow for \colorul{finance}{finance} and \colorul{law}{law}; orange for \colorul{healthcare}{healthcare}; red for \colorul{education}{education} and \colorul{social_work}{social work}; and dark red for \colorul{media}{media}, \colorul{journalism}{journalism}, and \colorul{design}{design}. The x-axis corresponds to the real-world distribution of each occupation group (see \ref{app:occupation_statistics}), ranging from 100\% female workers on the left to a 50\% (50\% male) balance in the middle, and finally to 0\% (100\% male) on the right. The grey vertical line marks occupations with minimal gender imbalance in the real world. The y-axis represents the gender distribution within the translated challenge set. An ideal translation would result in all markers aligning with the green horizontal line, indicating preserved original distribution as WinoMTDE is balanced in terms of gender and stereotypes.}
  \label{fig:prediction_dist}
\end{figure*}

The \textbf{accuracy} results, measuring how well a model preserves the original gender, range from 37.0\% to 95.8\%, showing significant performance differences across models and language pairs. GPT-4o-mini consistently achieves the highest accuracy across all language pairs, outperforming other MT systems. SYSTRAN, the only hybrid model evaluated, performs particularly well in the Romance language family, outperforming most neural models. Performance in the Slavic languages is weaker, with only GPT-4o-mini and DeepL surpassing the 50\% accuracy threshold, which indicates a performance better than random guessing. Google Translate performs the weakest overall.\\
The \textbf{gender-based F1-score gap} ($\Delta_G$) measures disparities between translations of male and female instances, with zero being the optimal score. Positive values indicate better performance on male instances, while a negative value reflects better performance on female instances. Across all models, the results reveal a consistent bias, with an average $\Delta_G$ of 11.9\%, indicating that models generally perform better on male instances. GPT-4o-mini stands out with the lowest average (2.4\%) across language pairs, indicating a less gender biased translation. DeepL also shows relatively good performance, even though it is outperformed by SYSTRAN in the Romance language family. \\
The \textbf{stereotype-based performance gap} ($\Delta_S$), which measures differences between stereotypical and anti-stereotypical translations, averages at 8.51\% across all models and languages. The performance gap is most noticeable in Romance languages, where Amazon Translate exhibits the largest $\Delta_S$, showing a strong bias towards non-stereotypical gender roles. Generally, GPT-4o-mini consistently exhibits lower scores than the other models, except for a strong non-stereotypical bias in Russian translations. 

The patterns observed in Table \ref{tab:results} are further illustrated in Figure \ref{fig:prediction_dist}, which visualizes the distribution of gender predictions across different occupational groups and models. While GPT-4o-mini often closely aligns, i.e. is within the green margin, with the perfect translation of the balanced WinoMTDE dataset, other models exhibit patterns that reflect or exaggerate real-world gender imbalances. Generally speaking, the results show that models tend to exhibit a strong bias towards male translations across all occupational groups, as indicated by the majority of markers falling into the upper two quadrants.

Overall, the results demonstrate that GPT-4o-mini achieves the strongest performance across all metrics, with SYSTRAN and DeepL also performing competitively, especially in Romance languages. The findings highlight significant weaknesses in Google Translate, which underperforms despite its widespread use.
\section{Discussion}
Generally, our results find that underrepresentation of females as well as stereotypical bias, although not as pronounced, is prevalent in most MT system. GPT-4o-mini, a large language model, consistently outperforms traditional MT systems, such as Google Translate, Microsoft Translator, Amazon Translate, and DeepL.
Nevertheless, it exhibits bias, particularly in Russian translations. This may stem from OpenAI's use of user data to train its models, but prohibiting Russians access to their models. This could lead to a lack of data and therefore the model might not be able to generalize well to the Russian language.
Furthermore, the results suggest that using hybrid MT models, such as SYSTRAN can lead to better results in Romance languages. This is especially interesting as it indicates that using set grammatical rules could be a possibility to minimize gender bias within MT from German to Romance languages. However,
SYSTRAN performed worse than the other MT models within the Slavic and Semitic language families. This could be due to the fact that the grammatical rules to translate to those languages are more complex and therefore harder to implement.

\subsection{Limitations and Future Work}
Despite providing valuable insights, this evaluation has several limitations. First, the WinoMTDE dataset is relatively small (288 sentences), potentially limiting the scope of gender bias that can be assessed. Stereotype annotations were based on German labor statistics and annotated by a single person, which may introduce bias, especially for ambiguous job titles (e.g., \textit{UntersucherIn}, meaning both ``examiner'' and ``investigator''). Additionally, the broad grouping of occupations fails to capture nuanced stereotypes within fields. The dataset also lacks non-binary pronouns and neutral job titles, restricting the analysis to a binary gender perspective and overlooking broader gender biases. Certain biases, like semantic derogation, are also unaddressed. For example, translating ``teacher'' into Spanish produced gendered terms (\textit{maestra} for female and \textit{profesor} for male subjects), reinforcing stereotypes.

\begin{figure}[h]
  \centering
  \includegraphics[width=\columnwidth]{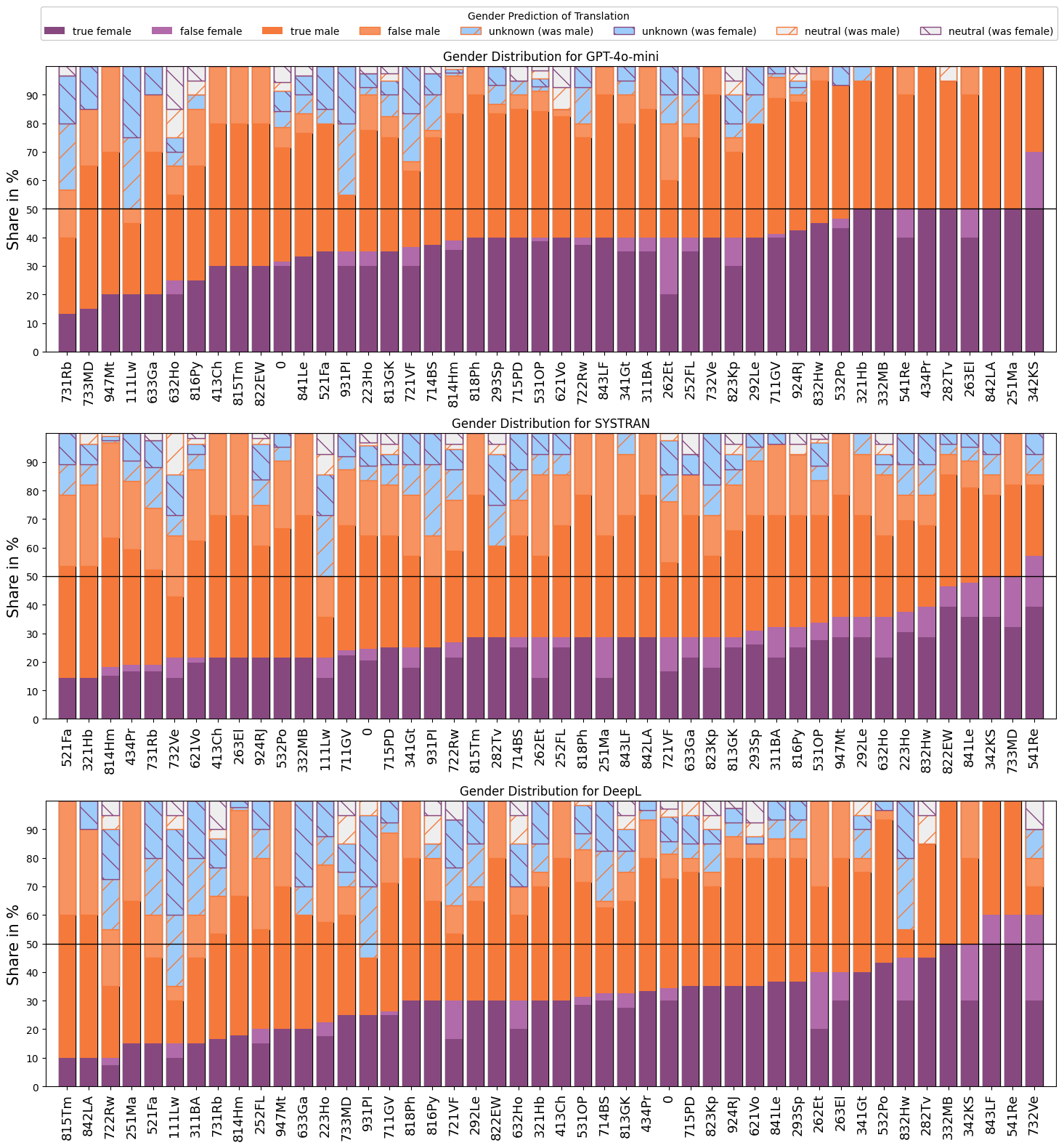}
  \caption[Gender Distribution of Translations by GPT-4o-mini, DeepL and SYSTRAN]{Depiction of the percentage of female (violet), male (orange), neutral (blue), and unknown (light blue) translations across occupations. Dark shades represent correct gender matches, light shades indicate errors. Hatching shows the gender origin within neutral and unknown categories. The horizontal line marks the 50/50 male-female ground truth.}
  \label{fig:romance}
\end{figure}

Moreover, this paper reports a higher share of unknown predictions compared to prior work, likely due to challenges with sentence alignment in \textit{fast-align}, particularly with complex German structures. SYSTRAN, a hybrid model, showed fewer unknown predictions, possibly due to its rule-based approach (see Figure \ref{fig:romance}). Thus, the models' actual accuracy might be higher than reported. A table of accuracies excluding unknown predictions is provided in the appendix (see \ref{app:accuracy_wo_unknown}).

Future work should address these limitations by expanding the dataset, refining stereotype annotations, and including non-binary pronouns and neutral job titles. The evaluation pipeline could also be improved by using more advanced alignment tools to reduce unknown predictions. Additionally, evaluating MT models with known architectures and training data may provide deeper insights into observed biases.
\subsection{Conclusion}
In order to better understand gender bias and its emerging harms,
it is crucial to evaluate MT systems systematically. The WinoMTDE
dataset and evaluation methodology provide the first foundation
for this evaluation in German MT. The results of the evaluation of five MT models and a general-purpose LLM, highlight the persistent gender bias within translations. These results emphasize the urgent need to develop more inclusive and equitable MT systems that ensure both accuracy and fairness in translations.
\newpage \newpage
\bibliography{main}

\begin{thebibliography}{18}
\providecommand{\natexlab}[1]{#1}

\bibitem[{Adler and Elhadad(2006)}]{adler_unsupervised_2006}
Meni Adler and Michael Elhadad. 2006.
\newblock \href {https://doi.org/10.3115/1220175.1220259} {An {Unsupervised}
  {Morpheme}-{Based} {HMM} for {Hebrew} {Morphological} {Disambiguation}}.
\newblock In \emph{Proceedings of the 21st {International} {Conference} on
  {Computational} {Linguistics} and 44th {Annual} {Meeting} of the
  {Association} for {Computational} {Linguistics}}, pages 665--672, Sydney,
  Australia. Association for Computational Linguistics.

\bibitem[{Blodgett et~al.(2020)Blodgett, Barocas, Daum{\'e}~III, and
  Wallach}]{blodgett-etal-2020-language}
Su~Lin Blodgett, Solon Barocas, Hal Daum{\'e}~III, and Hanna Wallach. 2020.
\newblock \href {https://doi.org/10.18653/v1/2020.acl-main.485} {Language
  (technology) is power: A critical survey of
  {\textquotedblleft}bias{\textquotedblright} in {NLP}}.
\newblock In \emph{Proceedings of the 58th Annual Meeting of the Association
  for Computational Linguistics}, pages 5454--5476, Online. Association for
  Computational Linguistics.

\bibitem[{Canfora and Ottmann(2020)}]{Canfora_Ottmann_2020}
Carmen Canfora and Angelika Ottmann. 2020.
\newblock \href {https://doi.org/10.1075/ts.00021.can} {Risks in neural machine
  translation}.
\newblock \emph{Translation Spaces}, 9(1):58–77.

\bibitem[{Chan and Tang(2024)}]{Chan_Tang_2024}
Venus Chan and William Ko-Wai Tang. 2024.
\newblock \href {https://doi.org/10.1007/s42979-024-03340-z} {Gpt for
  translation: A systematic literature review}.
\newblock \emph{SN Computer Science}, 5(8):986.

\bibitem[{Costa-jussà(2019)}]{costa-jussa_analysis_2019}
Marta~R. Costa-jussà. 2019.
\newblock \href {https://doi.org/10.1038/s42256-019-0105-5} {An analysis of
  gender bias studies in natural language processing}.
\newblock \emph{Nature Machine Intelligence}, 1(11):495--496.

\bibitem[{Dyer et~al.(2013)Dyer, Chahuneau, and Smith}]{dyer_simple_2013}
Chris Dyer, Victor Chahuneau, and Noah~A. Smith. 2013.
\newblock \href {https://aclanthology.org/N13-1073} {A {Simple}, {Fast}, and
  {Effective} {Reparameterization} of {IBM} {Model} 2}.
\newblock In \emph{Proceedings of the 2013 {Conference} of the {North}
  {American} {Chapter} of the {Association} for {Computational} {Linguistics}:
  {Human} {Language} {Technologies}}, pages 644--648, Atlanta, Georgia.
  Association for Computational Linguistics.

\bibitem[{Horvath et~al.(2016)Horvath, Merkel, Maass, and
  Sczesny}]{horvath_does_2016}
Lisa~K. Horvath, Elisa~F. Merkel, Anne Maass, and Sabine Sczesny. 2016.
\newblock \href {https://doi.org/10.3389/fpsyg.2015.02018} {Does
  {Gender}-{Fair} {Language} {Pay} {Off}? {The} {Social} {Perception} of
  {Professions} from a {Cross}-{Linguistic} {Perspective}}.
\newblock \emph{Frontiers in Psychology}, 6.

\bibitem[{Keep et~al.(2021)Keep, Oerlemans, Raes, Tresoor, and
  Wijnhoven}]{citekey}
Matthijs Keep, Jeroen Oerlemans, Rik Raes, Milan Tresoor, and Bert Wijnhoven.
  2021.
\newblock Evaluating gender bias in dutch machine translation.
\newblock Unpublished.

\bibitem[{Levesque et~al.(2012)Levesque, Davis, and
  Morgenstern}]{levesque_winograd_2012}
Hector~J. Levesque, Ernest Davis, and Leora Morgenstern. 2012.
\newblock The {Winograd} schema challenge.
\newblock In \emph{Proceedings of the {Thirteenth} {International} {Conference}
  on {Principles} of {Knowledge} {Representation} and {Reasoning}}, {KR}'12,
  pages 552--561, Rome, Italy. AAAI Press.

\bibitem[{Patil and Davies(2014)}]{patil_use_2014}
Sumant Patil and Patrick Davies. 2014.
\newblock \href {https://doi.org/10.1136/bmj.g7392} {Use of {Google}
  {Translate} in medical communication: evaluation of accuracy}.
\newblock \emph{BMJ}, 349:g7392.
\newblock Publisher: British Medical Journal Publishing Group Section:
  Research.

\bibitem[{Rudinger et~al.(2018)Rudinger, Naradowsky, Leonard, and
  Van~Durme}]{rudinger_gender_2018}
Rachel Rudinger, Jason Naradowsky, Brian Leonard, and Benjamin Van~Durme. 2018.
\newblock \href {http://arxiv.org/abs/1804.09301} {Gender {Bias} in
  {Coreference} {Resolution}}.
\newblock \emph{arXiv preprint}.
\newblock ArXiv:1804.09301 [cs].

\bibitem[{Savoldi et~al.(2021)Savoldi, Gaido, Bentivogli, Negri, and
  Turchi}]{savoldi_gender_2021}
Beatrice Savoldi, Marco Gaido, Luisa Bentivogli, Matteo Negri, and Marco
  Turchi. 2021.
\newblock \href {https://doi.org/10.1162/tacl_a_00401} {Gender {Bias} in
  {Machine} {Translation}}.
\newblock \emph{Transactions of the Association for Computational Linguistics},
  9:845--874.

\bibitem[{Stanovsky et~al.(2019)Stanovsky, Smith, and
  Zettlemoyer}]{stanovsky-etal-2019-evaluating}
Gabriel Stanovsky, Noah~A. Smith, and Luke Zettlemoyer. 2019.
\newblock \href {https://doi.org/10.18653/v1/P19-1164} {Evaluating gender bias
  in machine translation}.
\newblock In \emph{Proceedings of the 57th Annual Meeting of the Association
  for Computational Linguistics}, pages 1679--1684, Florence, Italy.
  Association for Computational Linguistics.

\bibitem[{{Statistik der Bundesagentur für
  Arbeit}(2020)}]{statistik_der_bundesagentur_fur_arbeit_klassifikation_2020}
{Statistik der Bundesagentur für Arbeit}. 2020.
\newblock \href
  {https://statistik.arbeitsagentur.de/DE/Navigation/Grundlagen/Klassifikationen/Klassifikation-der-Berufe/KldB2010-Fassung2020/Arbeitsmittel/Arbeitsmittel-Nav.html#faq_1614736}
  {Klassifikation der {Berufe} 2010 – überarbeitete {Fassung} 2020}.

\bibitem[{Sterling et~al.(2020)Sterling, Thompson, Wang, Kusimo, Gilmartin, and
  Sheppard}]{sterling_confidence_2020}
Adina~D. Sterling, Marissa~E. Thompson, Shiya Wang, Abisola Kusimo, Shannon
  Gilmartin, and Sheri Sheppard. 2020.
\newblock \href {https://doi.org/10.1073/pnas.2010269117} {The confidence gap
  predicts the gender pay gap among {STEM} graduates}.
\newblock \emph{Proceedings of the National Academy of Sciences of the United
  States of America}, 117(48):30303--30308.

\bibitem[{Vervecken and Hannover(2015)}]{vervecken_yes_2015}
Dries Vervecken and Bettina Hannover. 2015.
\newblock \href {https://doi.org/10.1027/1864-9335/a000229} {Yes {I} {Can}!
  {Effects} of {Gender} {Fair} {Job} {Descriptions} on {Children}'s
  {Perceptions} of {Job} {Status}, {Job} {Difficulty}, and {Vocational}
  {Self}-{Efficacy}}.
\newblock \emph{Social Psychology}, 46:76--92.

\bibitem[{Vieira et~al.(2021)Vieira, O’Hagan, and
  O’Sullivan}]{vieira_understanding_2021}
Lucas~Nunes Vieira, Minako O’Hagan, and Carol O’Sullivan. 2021.
\newblock \href {https://doi.org/10.1080/1369118X.2020.1776370} {Understanding
  the societal impacts of machine translation: a critical review of the
  literature on medical and legal use cases}.
\newblock \emph{Information, Communication \& Society}, 24(11):1515--1532.

\bibitem[{Zhao et~al.(2018)Zhao, Wang, Yatskar, Ordonez, and
  Chang}]{zhao_gender_2018}
Jieyu Zhao, Tianlu Wang, Mark Yatskar, Vicente Ordonez, and Kai-Wei Chang.
  2018.
\newblock \href {https://doi.org/10.18653/v1/N18-2003} {Gender {Bias} in
  {Coreference} {Resolution}: {Evaluation} and {Debiasing} {Methods}}.
\newblock In \emph{Proceedings of the 2018 {Conference} of the {North}
  {American} {Chapter} of the {Association} for {Computational} {Linguistics}:
  {Human} {Language} {Technologies}, {Volume} 2 ({Short} {Papers})}, pages
  15--20, New Orleans, Louisiana. Association for Computational Linguistics.

\end{thebibliography}

\appendix
\onecolumn
\section{Appendix}
\subsection{Occupation Statistics}\label{app:occupation_statistics}
\begin{table}[H]
    \centering
    \small
    \begin{tabular}{llp{0.45\textwidth}}
        \toprule
        \textbf{Code} & \textbf{Occupational Group Name} & \textbf{Job Instances} \\
        \midrule
        111Lw & Landwirtschaft & Landwirt, Landwirtin \\
        223Ho & Holzbe- und -verarbeitung & Schreiner, Schreinerin, Tischler, Tischlerin, Umzugshelferin, Umzugshelfer \\
        251Ma & Maschinenbau- und Betriebstechnik & Ingenieur, Maschinistin, Ingenieurin, Maschinist \\
        252FL & Fahrzeug-Luft-Raumfahrt-, Schiffbautechn. & Mechaniker, Mechanikerin \\
        262Et & Energietechnik & Elektriker, Elektrikerin \\
        263El & Elektrotechnik & Techniker, Technikerin \\
        282Tv & Textilverarbeitung & Schneider, Schneiderin \\
        292Le & Lebensmittel- u. Genussmittelherstellung & Bäcker, Bäckerin \\
        293Sp & Speisenzubereitung & Köchin, Koch, Chefkoch, Chefköchin \\
        311BA & Bauplanung u. -überwachung, Architektur & Architektin, Architekt, Planer, Planerin \\
        321Hb & Hochbau & Bauarbeiter, Bauarbeiterin \\
        332MB & Maler., Stuckat., Bauwerksabd., Bautenschutz & Malerin, Maler \\
        341Gt & Gebäudetechnik & Hausmeister, Hausmeisterin \\
        342KS & Klempnerei, Sanitär, Heizung, Klimatechnik & Klempnerin, Klempner \\
        413Ch & Chemie & Chemikerin, Chemiker \\
        434Pr & Softwareentwicklung und Programmierung & Entwicklerin, Entwickler, Programmierer, Programmiererin \\
        521Fa & Fahrzeugführung im Stra{\ss}enverkehr & Fahrerin, Fahrer \\
        531OP & Obj.-, Pers.-, Brandschutz, Arbeitssicherh. & Aufseher, Aufseherin, Wachfrau, Wachmann, Feuerwehrfrau, Feuerwehrmann, Ermittlerin, Inspektorin, Inspektor, Ermittler \\
        532Po & Polizei, Kriminald., Gerichts, Justizvollz. & Polizistin, Polizist, Polizisten \\
        541Re & Reinigung & Reiniger, Reinigerin \\
        621Vo & Verkauf (ohne Produktspezialisierung) & Kassierer, Kassiererin, Verkäuferin, Verkäufer \\
        632Ho & Hotellerie & Rezeptionistin, Rezeptionisten \\
        633Ga & Gastronomie & Barkeeper, Barkeeperin \\
        711GV & Geschäftsführung und Vorstand & Geschäftsführer, Geschäftsführerin, Chef, Chefin, Manager, Managerin, Vorgesetzte, Vorgesetzten \\
        714BS & Büro und Sekretariat & Assistent, Assistentin, Sekretärin, Sekretär \\
        715PD & Personalwesen und -dienstleistung & Beraterin, Berater \\
        721VF & Versicherungs- u. Finanzdienstleistungen & Analystin, Analyst, Aktienmaklerin, Aktienmakler \\
        722Rw & Rechnungswesen, Controlling und Revision & Buchhalter, Buchhalterin, Wirtschaftsprüfer, Wirtschaftsprüferin \\
        731Rb & Rechtsberatung, -sprechung und -ordnung & Anwältin, Anwalt, Rechtsassistent, Rechtsassistentin \\
        732Ve & Verwaltung & Verwalter, Verwalterin \\
        733MD & Medien-Dokumentations-Informationsdienst & Bibliothekar, Bibliothekarin \\
        813GK & Gesundh., Krankenpfl., Rettungsd., Geburtsh. & Krankenpflegerin, Krankenpfleger, Disponentin, Disponenten, Sanitäterin, Sanitäter \\
        814Hm & Human- und Zahnmedizin & Arzt, Ärztin, Fachärztin, Facharzt, Hausarzt, Hausärztin, Untersucherin, Untersucher, Hygienikerin, Hygieniker, Pathologin, Pathologe, Chirurgin, Chirurg \\
        815Tm & Tiermedizin und Tierheilkunde & Tierärztin, Tierarzt \\
        816Py & Psychologie, nichtärztl. Psychotherapie & Therapeutin, Psychologin, Therapeuten, Psychologe \\
        818Ph & Pharmazie & Apothekerin, Apotheker \\
        822EW & Ernährungs-, Gesundheitsberatung, Wellness & Ernährungsberater, Ernährungsberaterin \\
        823Kp & Körperpflege & Friseurin, Friseur \\
        832Hw & Hauswirtschaft und Verbraucherberatung & Haushälter, Haushälterin \\
        841Le & Lehrtätigkeit an allgemeinbild. Schulen & Lehrer, Lehrerin \\
        842LA & Lehrt. berufsb. Fächer, betr. Ausbildung, Betr. päd. & Instrukteurin, Instrukteur \\
        843LF & Lehr-, Forschungstätigkeit an Hochschulen & Wissenschaftlerin, Wissenschaftler \\
        924RJ & Redaktion und Journalismus & Redakteur, Redakteurin, Schriftsteller, Schriftstellerin \\
        931PI & Produkt- und Industriedesign & Designer, Designerin \\
        947Mt & Museumstechnik und -management & Gutachter, Gutachterin \\
        0 & Allgemein & Angestellter, Angestellte, Arbeiter, Arbeiterin, Mitarbeiter, Mitarbeiterin, Steuerzahler, Steuerzahlerin \\
        \bottomrule
    \end{tabular}
    \caption{Occupation Statistics of the German Department of Labor. All occupational groups present in the dataset are displayed. Code denotes the labeling of \textit{Statistik der Bundesagentur für Arbeit Klassifikation 2020}, with each occupation having a unique code for reference. Furthermore, all job instances from the WinoMTDE challenge set are namely displayed.}\label{tab:occupation_statistics}
\end{table}
\subsection{Accuracy Results without Unknown Predictions}\label{app:accuracy_wo_unknown}
\begin{table}[H]
        \resizebox{1\textwidth}{!}{%
        \begin{tabular}{ccccccccccccc}
        \toprule
        \multicolumn{1}{c}{} & \multicolumn{2}{c}{\textbf{GT}} &  \multicolumn{2}{c}{\textbf{MT}} & \multicolumn{2}{c}{\textbf{AT}} &  \multicolumn{2}{c}{\textbf{ST}} & \multicolumn{2}{c}{\textbf{DL}} & \multicolumn{2}{c}{\textbf{4o-min}} \\
        \cmidrule(lr){2-3} \cmidrule(lr){4-5} \cmidrule(lr){6-7} \cmidrule(lr){8-9} \cmidrule(lr){10-11} \cmidrule(lr){12-13}
        \textbf{Languages} & \textsc{Acc}' & \textsc{Acc} & \textsc{Acc}' & \textsc{Acc} & \textsc{Acc}' & \textsc{Acc} & \textsc{Acc}' & \textsc{Acc} & \textsc{Acc}' & \textsc{Acc} & \textsc{Acc}' & \textsc{Acc} \\
        \midrule
        \textit{DE$\rightarrow$ES} & 78.0 & 66.8 & 72.7 & 62.0 & 83.2 & 72.7 & 96.8 & 94.1 & 89.7 & 83.1 & 99.0 & 95.8 \\
        \textit{DE$\rightarrow$FR} & 73.0 & 64.2 & 77.8 & 69.2 & 76.2 & 68.0 & 87.0 & 80.6 & 92.5 & 83.3 & 94.9 & 89.4 \\
        \textit{DE$\rightarrow$IT} & 65.8 & 52.0 & 63.8 & 51.8 & 76.2 & 58.9 & 82.2 & 70.9 & 81.2 & 61.9 & 87.4 & 75.5 \\
        \midrule
        \textit{DE$\rightarrow$UK} & 66.4 & 46.5 & 62.1 & 48.2 & 62.5 & 41.4 & 49.4 & 38.2 & 68.2 & 54.7 & 80.7 & 69.6 \\
        \textit{DE$\rightarrow$RU} & 60.4 & 42.7 & 63.3 & 46.4 & 62.2 & 47.3 & 52.8 & 37.0 & 60.6 & 42.3 & 66.5 & 55.4 \\
        \midrule
        \textit{DE$\rightarrow$AR} & 67.4 & 55.2 & 66.0 & 54.0 & 68.2 & 59.2 & 62.0 & 51.5 & - & - & 88.9 & 83.3 \\
        \textit{DE$\rightarrow$HE} & 72.9 & 64.5 & 77.2 & 65.4 & 69.4 & 60.3 & 52.1 & 44.6 & - & - & 84.0 & 78.1 \\
        \bottomrule
        \end{tabular}}
        \caption[Accuracy Results without Unknown Gender Predictions]{Accuracy results without unknown gender predictions. For each MT model and all languages grouped within their respective family, the accuracy is provided. The first column displays \textsc{Acc}', denoting the accuracy values that do not include unknown predictions, and the second column displays the accuracy presented previously, including all predictions.}
\end{table}
\twocolumn

\end{document}